\documentclass[journal]{IEEEtran}
\usepackage{setspace}
\usepackage{cite}
\usepackage{balance}

\ifCLASSINFOpdf

\else

\fi

\usepackage{graphicx}
\usepackage{amsmath}
\usepackage{amssymb}
\usepackage{multicol}
\usepackage{verbatim}
\usepackage{booktabs}
\usepackage{multirow}
\usepackage{siunitx}
\usepackage{romannum}
\usepackage{hyperref}
\hypersetup{colorlinks,citecolor={blue}}  
\usepackage{float}
\usepackage{caption}

\usepackage[utf8]{inputenc} 
\usepackage[T1]{fontenc}    
\usepackage{url}            
\usepackage{booktabs}       
\usepackage{amsfonts}       
\usepackage{nicefrac}       
\usepackage{doi}
\usepackage{tabularx,booktabs}
\usepackage{graphics}
\usepackage{svg}
\usepackage{multirow}
\usepackage[export]{adjustbox}
\usepackage{algpseudocode}
\usepackage{algorithm}
\usepackage{esvect}
\usepackage{scalerel}
\usepackage{fixltx2e}

\usepackage[left=2cm,right=2cm,top=2cm,bottom=2cm]{geometry}
\usepackage{url}
\usepackage{xr-hyper}

\makeatletter
\newcommand*{\addFileDependency}[1]{
  \typeout{(#1)}
  \@addtofilelist{#1}
  \IfFileExists{#1}{}{\typeout{No file #1.}}
}
\makeatother

\newcommand*{\myexternaldocument}[1]{%
    \externaldocument{#1}%
    \addFileDependency{#1.tex}%
    \addFileDependency{#1.aux}%
}

\myexternaldocument{supporting_documents}

\begin{document}

\title{Unsupervised Simplification of Legal Texts}

\author{Mert Cemri, Tolga \c{C}ukur*,~\IEEEmembership{Senior Member,~IEEE}, and Aykut Ko\c{c}*,~\IEEEmembership{Senior Member,~IEEE}
\thanks{
M. Cemri, T. \c{C}ukur, and A. Ko\c{c} are with the Dept. of Electrical and Electronics Engineering, Bilkent University, TR-06800 Bilkent, Ankara, Turkey, and also with UMRAM, Bilkent University, TR-06800 Ankara, Turkey. (e-mails: mert.cemri@ug.bilkent.edu.tr, cukur@ee.bilkent.edu.tr, aykut.koc@bilkent.edu.tr.)  *Co-Senior Authors.
}
}

\markboth{}%
{Shell \MakeLowercase{\textit{et al.}}: Bare Demo of IEEEtran.cls for IEEE Journals}




\maketitle

\begin{abstract}
The processing of legal texts has been developing as an emerging field in natural language processing (NLP). Legal texts contain unique jargon and complex linguistic attributes in vocabulary, semantics, syntax, and morphology. Therefore, the development of text simplification (TS) methods specific to the legal domain is of paramount importance for facilitating comprehension of legal text by ordinary people and providing inputs to high-level models for mainstream legal NLP applications. While a recent study proposed a rule-based TS method for legal text, learning-based TS in the legal domain has not been considered previously. Here we introduce an unsupervised simplification method for legal texts (USLT). USLT performs domain-specific TS by replacing complex words and splitting long sentences. To this end, USLT detects complex words in a sentence, generates candidates via a masked-transformer model, and selects a candidate for substitution based on a rank score. Afterward, USLT recursively decomposes long sentences into a hierarchy of shorter core and context sentences while preserving semantic meaning. We demonstrate that USLT outperforms state-of-the-art domain-general TS methods in text simplicity while keeping the semantics intact. 
\end{abstract}


\pagenumbering{arabic}

\IEEEpeerreviewmaketitle
\section{Introduction}
\label{introduction}

\IEEEPARstart{L}{aw} and judiciary systems constitute one of the essential pillars of a functioning society. Heavily dependent on written documents, law is a prime application domain for NLP \cite{dozier10ner,aleven03cato,ashley09automatically,chalkidis19deep}. Legal professionals must perform the arduous tasks of comprehending, analyzing, and processing legal documents. Primarily based on NLP techniques, there is an explosion of interest in developing algorithms for high-level legal technology applications \cite{zhong20summary,chalkidis2021lexglue,legalbert,galgani12nlpInLaw,chalkidis19classification}. These include legal judgement forecasting \cite{ruger04decisionTreePred,aletras16svmPredicting,sulea17svmPredicting,katz17randomForestPred,long19automatic,chalkidis19neural,kaufman19improving,osullivan19predicting,medvedeva20caseData,yang19multiPerspective,strickson20ljpUK,MUMCUOGLU2021102684,canadian_judgement}, legal topic classification \cite{oneill17sententialModality,howe19signapore, chalkidis19classification}, legal text summarization \cite{galgani12nlpInLaw, kanapala19legalTextSummarization}, legal question\&answer systems \cite{kien20answering,kim17applying,ravichander19question,zhong20iteratively,vietnamese_law}, gender debiasing in legal corpora \cite{sevim_sahinuc_koc_2022}, information and feature extraction from legal contracts and documents \cite{legal_correspondence,nguyen18recurrent,9648744,8934044}, legal named-entity recognition (legal-NER) \cite{cardellino17legalNerC,dozier10ner,elnaggar18nerTransfer,cetindag_yazicioglu_koc_2022}, and court opinion generation \cite{ye18courtView}. Following the advancements in machine learning and NLP-based computational law, standardized benchmarks have also emerged \cite{chalkidis2021lexglue}. The long-standing relationship between AI and Law has been comprehensively surveyed in recent publications \cite{thirty_years_editor,thirty_years_third_decade}.

The legal field has a domain-specific sub-language with complex jargon, called Legal English, or ``Legalese" \cite{simpatico}. Legal English differs from its regular counterpart in terms of vocabulary related to technical terminology, morphology, semantics, and other specific features such as the use of uncommon pro-forms and word order. While employed by lawyers to bring precision and clarity to written documents, Legal English is naturally difficult to understand for ordinary people. Therefore, simplifying legal texts is essential to improve the comprehensibility of legal documents by the general public and legal professionals. One reason that legal texts are hard to grasp is the usage of archaic and overly complicated vocabulary, such as \verb|witnesseth|, \verb|hereinafter|, \verb|injunction|; as well as Latin and French phrases such as \verb|mens rea| (criminal intent), \verb|prima facie| (at first appearance) and \verb|actus reus| (guilty act). Another reason is that legal texts contain prolonged sentences \cite{legal_text_simplification_impact}, with multiple clauses \cite{simpatico}.

Text simplification (TS) methods offer a promising solution to this problem by transforming a complex piece of text into an easier-to-understand form. Domain-general TS studies have reported considerable benefits of simplification in regular text, especially for disadvantaged populations including children \cite{lsforchildren}, people with genetic disorders such as dyslexia and autism \cite{Carroll1999SimplifyingTF,lsforaphasic}, and for language learners \cite{lsfornonnatives}. However, due to the divergent characteristics of legal language from everyday language, domain-general methods can function sub-optimally on legal texts. To address this issue, a recent study has introduced a rule-based TS method specifically devised for the legal domain \cite{simpatico}. However, non-learning-based methods leverage hand-constructed features that can elicit suboptimal generalization performance.

Here we introduce an unsupervised learning-based TS method, USLT, to mold legal texts into a more accessible form for the general public. To simplify complex vocabulary, USLT performs lexical simplification (LS) by finding appropriate replacements for complex lexical units in a sentence while retaining the semantics and grammar. To identify complex units, a method specialized for legal texts is proposed that characterizes word complexity in terms of frequencies of occurrence in legal versus general corpora \cite{lsforchildren,lsforaphasic,simpatico}. A domain-specific language model (LM), Legal-BERT \cite{legalbert}, is then used to generate substitution candidates for complex units, as inspired by recent domain-general LS studies \cite{qiang21lsbert}. To select a substitute, candidates are ranked according to word features, including word length, word frequency, LM-loss of the word in the sentence, the original predicted probability of the candidate by the LM, and the cosine similarity of the candidate word to the original complex word. USLT leverages a splitting module to process the lexically simplified sentences to address overly long sentences. In particular, a recursive algorithm is employed where long sentences are hierarchically split into core and surrounding context sentences while preserving semantic relationships.

We demonstrated USLT on two datasets: a legal corpus of 27,000 US Supreme Court cases \cite{supremecourtcases} and a legal corpus of 1,000 sentences from diverse case categories within the Case-Law Project \cite{caselaw}. USLT was compared against state-of-the-art domain-general unsupervised TS methods for regular text. The comparison methods included LS-BERT and REC-LS, which primarily perform lexical simplification \cite{qiang21lsbert,gooding-kochmar-2019-recursive}, and ACCESS and MUSS, which perform both lexical and syntactic simplification \cite{martin2020controllable,martin2021muss}. USLT outperforms all comparison methods in quantitative metrics of simplicity and performs on par with baselines in terms of the semantic coherence of the resultant text.

\section{Related Work}
\label{relworks}
TS aims to simplify a given sentence to make it more understandable to readers via lexical and/or syntactic alterations. Earlier TS methods were inspired by statistical machine translation, where the goal was to learn the simplification rules as a translation of a complex sentence into a simple sentence \cite{coster-kauchak-2011-simple}. Later studies adopted neural machine translation \cite{neural_machine_translation} based on architectures such as encoder-decoder models \cite{wang2016experimental,dong-etal-2019-editnts,nisioi-etal-2017-exploring}. Translation typically employs supervised models trained on large amounts of paired simple-complex sentences. Although few paired datasets are available for common language \cite{zhu-etal-2010-monolingual}, no such dataset exists for legal language to the best of our knowledge. Thus, unsupervised approaches are direly needed to improve the applicability of learning-based TS to legal texts. 

Several unsupervised approaches have been proposed for domain-general TS to date. \cite{lsfornonnatives} proposed to replace complex words in a sentence with simpler synonyms extracted from a thesaurus. \cite{surya-etal-2019-unsupervised,yatskar-etal-2010-sake} proposed unpaired training on separate corpora of simple sentences (Simple Wikipedia) and complex sentences (Wikipedia). Similarly, \cite{martin2021muss} proposed using data mining to extract simple and complex sentences from a single corpus to create paired datasets for supervised training. However, the abovementioned methods were devised for regular text, and paired corpora of simple and complex legal texts are rare. We introduce an unsupervised domain-specific LS method to address these limitations, followed by unsupervised sentence splitting for further structural simplification.

Developing domain-specific methods is demonstratedly important in many NLP applications, including simplification tasks \cite{sun2019demo,domain-specific-biomedical,zhou19influence,5560656}. 
Domain-specific TS methods have been proposed in medicine \cite{domain-specific-biomedical,8969670,7776380,kandula2010semantic} and finance \cite{maia2018finsslx}. Although domain-specific TS methods would have important implications in legal NLP \cite{legal_text_simplification_impact}, this problem remains largely unexplored. Only a recent study has introduced the SIMPATICO method that aims to simplify the Philippine Senate and House bills \cite{simpatico}. SIMPATICO uses a thesaurus to fetch candidates for simpler synonyms of complex words and a rule-based algorithm to select among candidates. Ruled-based LS methods define static rules to map each complex word to simpler synonyms without the need for a training dataset \cite{maddela-xu-2018-word}. However, they heavily rely on linguistic databases such as WordNet to fetch a predefined list of complex words and determine the simplest among these alternatives based on statistical metrics such as frequency of occurrence or character lengths of words \cite{lsforchildren,lsforaphasic}.

A powerful alternative is to use learning-based models to develop simplification methods. These methods include utilizing n-gram LMs to return the most likely substitutions for a word given its context \cite{ngram_LS}; learning rules from paired corpora \cite{horn-etal-2014-learning} and utilizing word embeddings to compute cosine similarity between an alternative word and a target complex word to find the most suitable replacements \cite{glavas-stajner-2015-simplifying,gooding-kochmar-2019-recursive,Qiang2021UnsupervisedST}. Recently, transformer-based models like Bidirectional Encoder Representation Transformers (BERT) \cite{devlin19bert}, have been successfully adopted in various NLP tasks \cite{wang21improving,guo21adaptive,wang20sbert,9739828}. Leveraging the masked language modeling (MLM) scheme in BERT \cite{chinese_masking,hubert,9721159}, substitution candidates can be generated for a masked word in a sentence. A seminal study for domain-general TS introduced the LS-BERT method based on BERT, and MLM \cite{qiang21lsbert}. Inspired by the success of this domain-general method, here we propose to leverage MLM on a legal-specific LM model, Legal-BERT \cite{legalbert}. Thus, our proposed method, USLT, aims to maximize domain-specific performance in simplifying legal texts while alleviating the need for paired training data. To our knowledge, USLT is the first learning-based TS method for legal texts.

The second important component of USLT is splitting long sentences into multiple shorter ones to facilitate comprehension. Previous studies on syntactic simplification used syntax-based hand-crafted transformation rules to do structural simplification operations \cite{ss11,ss12}. In contrast, others approached the problem using a semantic parser to partition a sentence into its primary semantic constitutes \cite{ss31,ss32}. Learning-based methods were also proposed, where a model is trained to learn simplification rules from samples of paired complex and simplified sentences \cite{ss21,ss22}. However, in the sentence splitting task, carefully hand-crafted rules produce relatively shorter and simpler sentences \cite{sentence_splitting}. Hence, we adapt the state-of-the-art sentence splitting method presented in \cite{sentence_splitting} to enable syntactic simplification without needing a paired dataset.

\section{Unsupervised Simplification of Legal Texts}

To simplify legal texts, USLT first performs lexical simplification on original sentences and then splits long sentences into shorter, more understandable sentences.

\subsection{Legal Language Model}
\label{bert_model_theory}

We first describe the details of the domain-specific LM that USLT leverages, namely Legal-BERT \cite{legalbert}. Legal-BERT is a domain-specific variant of BERT \cite{devlin19bert}, which is a transformer-based LM \cite{attention}.
BERT processes a sentence as a sequence of tokens. If a word is masked out in a sentence via annotation with the "[MASK]" token, MLM produces a probability distribution over its vocabulary, trying to find the most suitable fit for the masked word. This can be expressed as:
\begin{equation}
\label{eqn:BERT_equation_2}
P(\cdot|S\symbol{92}{w}) = P(\cdot|S'),
\end{equation}
where $S$ is the input sentence, $S\symbol{92}{w}=S'$ is the sentence with word $w$ masked. Therefore, if we were to mask complex words in a sentence, MLM can be used to search for proper substitution candidates according to the LM. In this study, as suggested by \cite{qiang21lsbert}, we proposed concatenating masked and unmasked versions of the sentence before feeding them to the LM, as $SS'$. The resultant conditional probability distribution over the vocabulary is given as:
\begin{equation}
\label{eqn:BERT_equation_with_double_sentences}
P(\cdot|S,S').
\end{equation}
This approach has three main advantages. First, since Legal-BERT model is trained using MLM, the model is adept at predicting suitable candidates for masked words. Second, instead of masking a single complex word at a time as proposed in \cite{gooding-kochmar-2019-recursive,qiang21lsbert}, we mask all complex words in $S$ simultaneously to prevent biases in attributed importance to any particular complex words. Third, providing the original unmasked sentence can help avoid the potential loss of contextual congruence during masking multiple complex words.

\subsection{Lexical Simplification Stage}

\label{three_step_section}

\begin{figure}[h]

\centering
{\includegraphics[width=0.8\columnwidth]{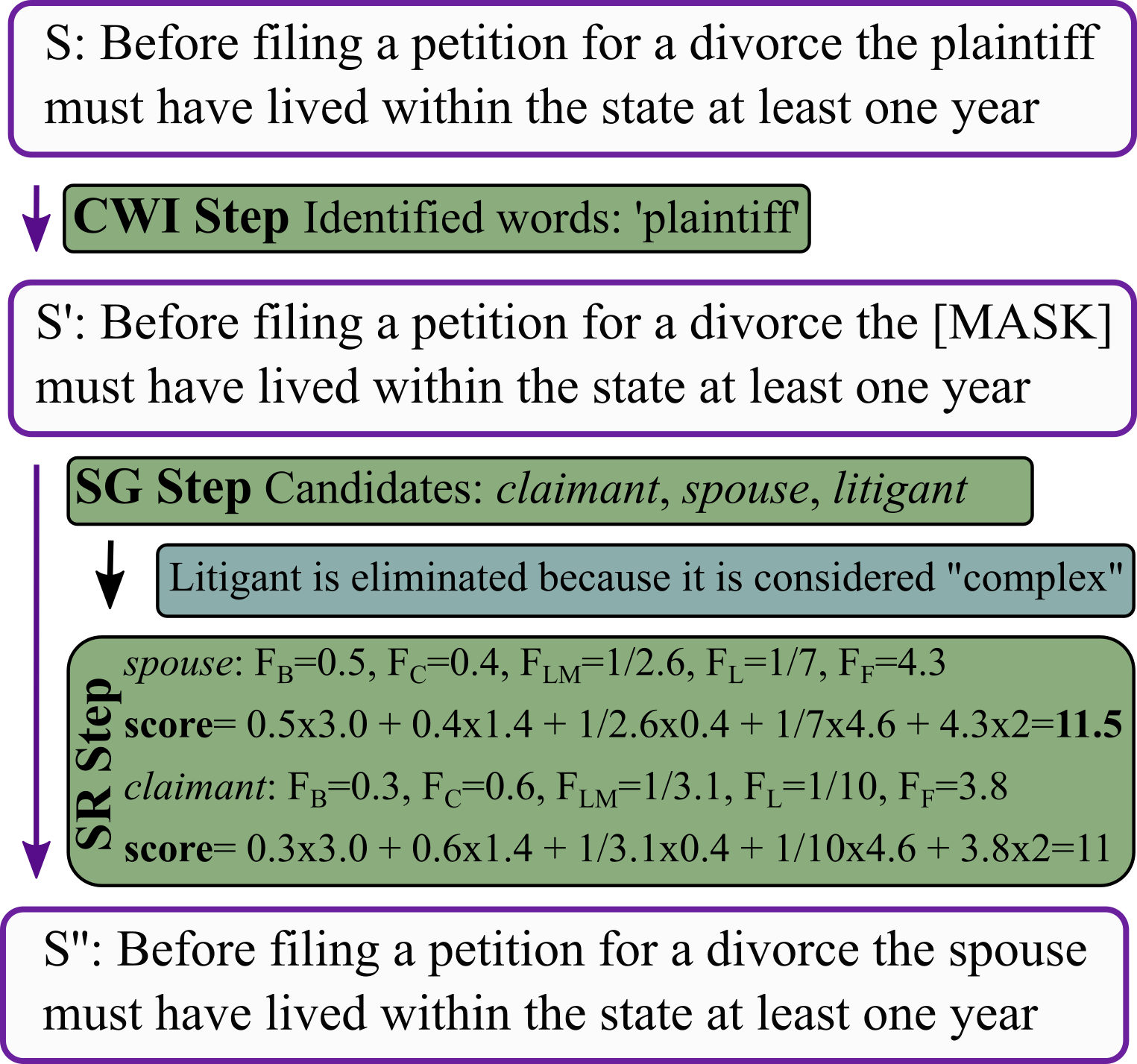}}
\captionsetup{justification=justified}
\caption{Schematic of the LS stage. First, on a given sentence $S$, CWI is carried out, and the identified complex words are masked to form $S'$. Then, $S'$ is fed into the LM for SG. The candidates surviving the eliminating steps are ranked using their features. The candidate with the highest total score is selected, and the simplified sentence, $S''$ is formed.}
\label{LS_figure}
\end{figure}

For a given complex sentence, the LS stage contains the following steps. In the complex word identification (CWI) step, complex words in the sentence are individually detected. The substitution generation (SG) step generates replacement candidates for those complex words. Next, candidates are ranked according to their word features in the substitution ranking (SR) step. Finally, identified complex words are replaced with the highest ranking candidate. Therefore, the LS stage in USLT comprises CWI, SG, and SR tasks, with the overall outline illustrated in Fig. \ref{LS_figure}.  

\subsubsection{Complex Word Identification (CWI)}
\label{complex_word_identification}

To handle the complex word identification task, we utilize the frequency of occurrences of words in two different corpora. Some recent studies approach this problem as a sequence labeling task or use convolutional neural networks (CNNs) \cite{aroyehun-etal-2018-complex,coster-kauchak-2011-simple}. However, our approach does not require a predetermined list of labeled complex words for model training. 

The first corpus is the SUBTLEX-UK corpus \cite{subtlexuk}. SUBTLEX-UK is a corpus containing English subtitles for movies in foreign languages, representing regular English in daily use. All words in SUBTLEX-UK are given a value from 1 to 7 for their frequencies of occurrence. This value, called Zipf-scale \cite{subtlexuk}, is inspired by the Zipf's Law of Word Frequencies, which is one of the most important statistical regularities of natural languages. Zipf's Law states that the frequency of a word is inversely proportional to its frequency rank. Furthermore, \cite{subtlexuk} provides a formula that enables assigning a number between 1 to 7 to each distinct word in a corpus (the higher the score, the more frequent the word). Therefore, the Zipf scale divides the frequency spectrum into seven discrete classes on a logarithmic scale. Examples of words corresponding to these discrete classes can be found in Table \ref{zipf_table_in_paper}.

\begin{table}[h]
\centering

\begin{tabular}{ |p{0.6cm}|p{0.7cm}||p{2.7cm}|p{3cm}| }

 \hline
 \textbf{Zipf Value} & \textbf{fpmw} & \textbf{SUBTLEX Corpus} & \textbf{Legal Corpus} \\
 \hline
 1 & 0.01 & antifungual, farsighted & exorbitantly, appelbaum \\
 2 & 0.1 & airstream, doorkeeper & unconscientious, abdicate\\
 3 & 1 & beanstalk, cornerstone  & manpower, cumbersome\\
 4 & 10  & fantasy, muffin  & interfering, violative\\
 5 & 100 & bedroom, drive & requirement, immunity\\
 6 & 1000 & day, great, other & judgement, district, power\\
 7 & 10000 & and, for, have & and, the, of \\
 \hline
\end{tabular}
\captionsetup{justification=justified}
\caption{Zipf values between 1 to 7, and the frequency per million words (fpmw), with some examples from the SUBTLEX and the legal corpora.}
\label{zipf_table_in_paper}
\end{table}

The second corpus comprises 27,000 legal cases retrieved from the US Supreme Court. We assigned Zipf values to each word as follows \cite{subtlexuk}:
\begin{align*}
  Zipf &= log_{10}(\frac{F + 1}{W + N}) + 3.0,
\end{align*}
where $F$ is the frequency of occurrence of a particular word, $W$ is the total number of words in the corpus in millions, and $N$ is the word types in the frequency list in millions. 

\begin{figure}[h]

\centering{\includegraphics[width=0.9\columnwidth]{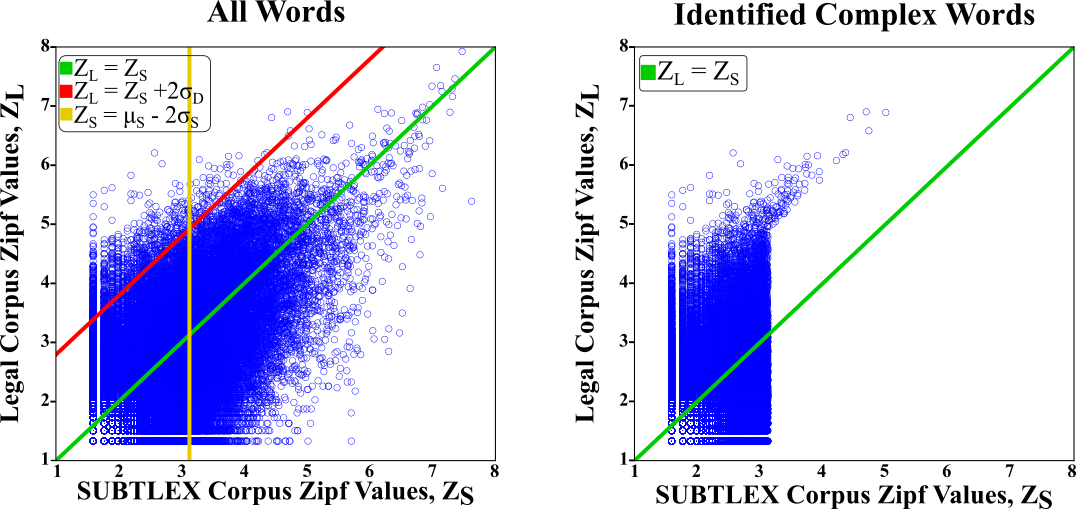}}

\captionsetup{justification=justified}
\caption{Plots of Zipf values of the words common in both SUBTLEX and legal corpora.}
\label{images_zipf}
\end{figure}

In order to identify complex words, we use two criteria. First, a word is considered complex in everyday language if it infrequently appears in the regular corpus. An upper bound on the Zipf value is defined to do this: 
\[
Z_S < \mu_S - 2\sigma_S,
\]
where $Z_S$ is the Zipf value of a particular word in SUBTLEX-UK corpus; $\mu_S$ and $\sigma_S$ are the mean and standard deviation of the Zipf values of the words in the SUBTLEX-UK corpus. Second, we consider words with domain-specific meanings that may need to be explained in simpler terms. To detect such words, we aim to identify words that are used much more commonly in the legal corpus than in the regular corpus. To this end, we consider a word ``complex" if it satisfies the following inequality:
\[
Z_S  < Z_L - 2\sigma_D,
\]
where $Z_L$ is the Zipf value of a particular word in the legal corpus, and $\sigma_D$ is the standard deviation of Zipf values of words in the legal corpus. This CWI procedure and the identified words are illustrated in Fig. \ref{images_zipf}.

Finally, for multi-word technical expressions in legal texts, such as ``actus reus" and ``ex gratia", we append a 400-word list \cite{legalphrases} to the list of complex words generated with the methods explained above.

\subsubsection{Substitution Generation (SG)}
\label{substitution_candidate_generation}

We check whether the word is a named-entity for each word in the given sentence. If it is not, we check whether that word is in the complex words list as defined in Section \ref{complex_word_identification}. If it is a complex word, then a substitution candidate is generated. For this purpose, all complex words in the sentence $S$ is masked to create the sentence $S'$. As explained in Section \ref{bert_model_theory}, the legal LM is then used to estimate the probability distribution over the vocabulary, and the top $n$ most likely tokens are retrieved.

\begin{figure}[h]
\centering{\includegraphics[width=1.0\columnwidth]{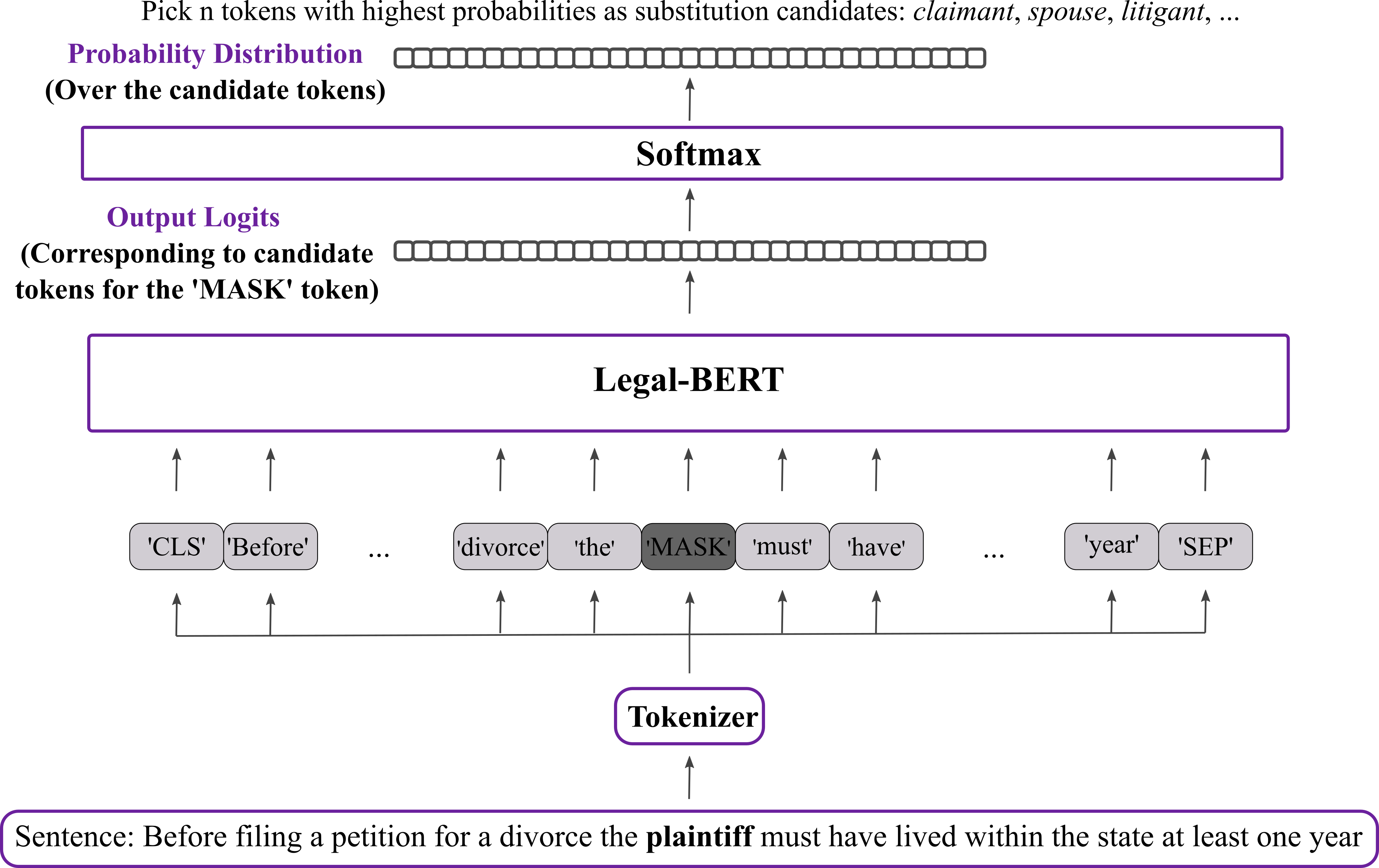}}
\captionsetup{justification=justified}
\caption{Generating substitution candidates. After masking the complex word and tokenizing the sentence, the tokens are fed into Legal-BERT to generate a list of candidate tokens, and tokens with the highest probabilities are retained.}
\label{bert_model}
\end{figure}

\subsubsection{Substitution Ranking (SR)}
\label{substition_candidate_ranking}

We first implement three elimination steps to rule out improper substitutions among the generated candidates and then perform numerical ranking among non-eliminated candidates. The elimination steps are the following:

\begin{itemize}
\item First, we do not want our targeted complex word to be replaced with another complex word. Hence, we remove candidates that are in the complex word list.
\item Second, we ensure that the candidates are real, meaningful words. Note that BERT models can occasionally generate sub-word tokens such as “\#\#ed”. To prevent this issue, we cross-check that each candidate appears in a broad list of 400,000 words, including nearly all common words in English \cite{vertanen_mobiletext}.
\item Third, part-of-speech (PoS) tags of candidates are checked to ensure that only candidate words with matching PoS attribute to the original word are retained. To implement this procedure, we leveraged the functionality offered by the nltk library \cite{nltk}.
\end{itemize}

Next, non-eliminated candidates are subjected to a scoring procedure to rank them. Various features of lexical units were proposed in the literature for this purpose \cite{qiang21lsbert,lsforaphasic,lsforchildren,lssurvey}. Since our proposed method is unsupervised, we avoid using a manual ranking system but develop a data-driven ranking score instead. In particular, we use a set of five weighted features to compute a ranking score. The weights are learned on a validation set (please see Section \ref{hyperparam_section}). 

The first feature, $F_B$, is the likelihood of the tokens produced by the LM for substitution. 

The second feature, $F_C$, reflects the semantic proximity of the candidate to the original word. To this end, the cosine similarity between the GloVe word embeddings of the original word $w$ and the candidate word $w'$ is used: 
\[ 
F_C = \frac{w \cdot w'}{\stretchleftright{\lVert}{w}{\rVert} \times
  \stretchleftright{\lVert}{w'}{\rVert}}.
\]

The third feature, $F_{LM}$, is the cross entropy loss calculated for the words near the candidate word. Precisely, we select a window of $5$ words centering the complex word, $(w_{-2},w_{-1},w_0,w_1,w_2)$, mask them one at a time, and compute the average cross-entropy loss over all 4 words:
\begin{align*}
  F_{LM} &= \frac{1}{L(w_{-2})+L(w_{-1})+L(w_{1})+L(w_{2})}.
\end{align*}

The fourth feature, $F_F$, is the frequency of the candidate word, where we take the Zipf values of the candidate word according to the SUBTLEX corpus as $F_F = Z_S$.

The fifth feature, $F_L$, is taken as the character length of a word. Numerous studies have highlighted or used the inverse proportionality between word length and word simplicity \cite{johannsen2012emnlp,Keskisarkka560901,lsforchildren}. To provide a theoretical ground for this, we use Zipf's Law of Abbreviation (also known as the Brevity Law), which states that longer words tend to be used less frequently, whereas shorter words are likely to be used frequently. Particularly, \cite{brevity_law} suggests that the quantitative analysis of the Zipf's Law of Abbreviation leads to the probability mass function of the frequency $n$ of a word, conditioned on length $\ell$ of that word to be proportional to a power of the word length as the following:
\[ 
f(n|\ell) \propto \frac{\ell^{\alpha_1 \delta}}{\ell^{\alpha_2 \delta}n^{\beta}},
\]
where $\alpha$ represents the power decay of the word frequency on a particular corpus. It is suggested to be $1.4$ if the word belongs to a corpus with a highly complex domain, such as the legal domain, and $2.75$ if it belongs to a domain with shorter words on average. $\delta$ is suggested to perfectly fit the Brevity Law when its value is $2.8$. Hence, we take  $\delta = 2.8, \alpha_1 = 1.4, \alpha_2 = 2.75$, resulting in the length feature:
\[ 
F_L = \ell^{-3.78},
\]
where $\ell$ is the number of letters in the candidate word.

To develop a data-driven aggregated score $sc$ based on the five features, a weighted combination is considered: \[sc =  W_B\times F_{B} + W_C\times F_{C}+ W_{LM}\times F_{LM}+W_{L}\times F_{L}+W_F\times F_{F}.\]
We proposed to learn the weights via Bayesian optimization \cite{bayesian_opt} as explained in Section \ref{hyperparam_section}, rank the candidates according to aggregate score, and finally select the highest ranking candidate as the replacement word.

\subsection{Sentence Splitting Stage}

After we carry out the LS, we perform hierarchical sentence splitting on the simplified albeit prolonged sentences. To this end, we adopt the algorithm described in \cite{sentence_splitting}, where a recursive approach based on hand-crafted rules is used to avoid the need for training datasets. In particular, long sentences are segregated into multiple shorter sentences while closely retaining the original sentence's grammatical structure and overall meaning. Therefore, we can make syntactic simplifications in addition to lexical changes in the given sentence. Hence, we can tackle the unusually lengthy structure of legal sentences, which is a primary reason that renders the sentences in legal documents hard to understand \cite{legal_text_simplification_impact}.

\begin{figure}[h]
\centering{\includegraphics[width=\columnwidth]{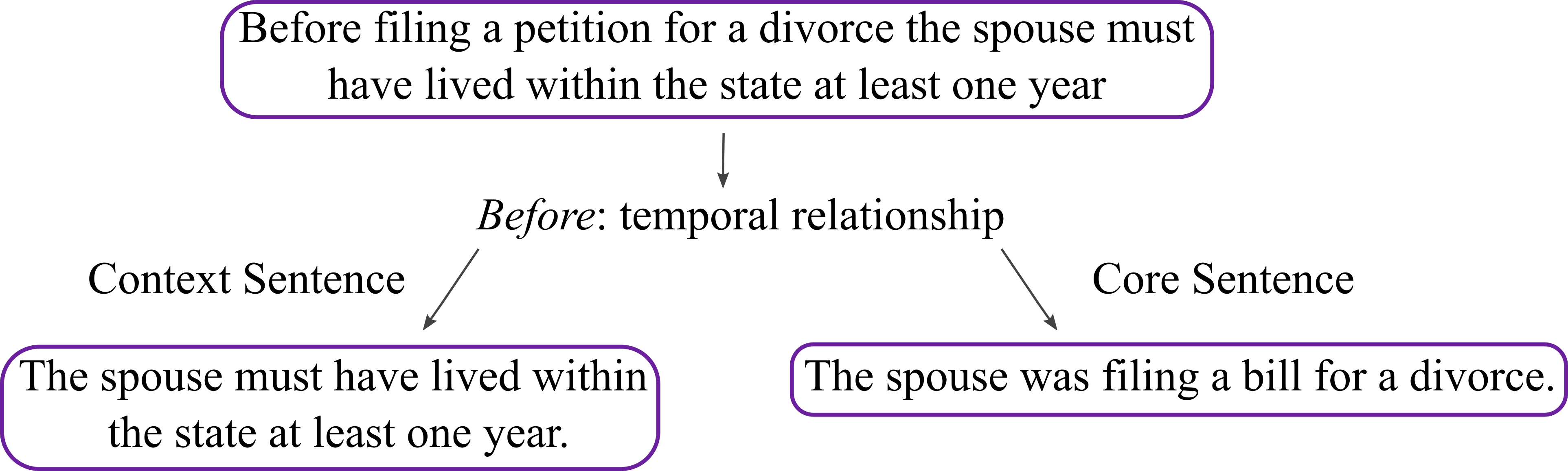}}
\caption{Sentence splitting of a lexically simplified sentence in Fig. \ref{LS_figure}. After the input sentence is split into a pair of structurally composed sentences, the subordinate clause, including the temporal relationship word \emph{before} is labeled as the context sentence. The other clause containing the input sentence's key information is labeled as the core sentence.}
\captionsetup{justification=justified}
\end{figure}

\section{Experimental Setup}
\label{experiments}

\subsection{Datasets}
\label{datasets}
In our experiments, we used two datasets for two different purposes. First, to optimize the substitution candidate ranking weights, we created a dataset of 500 sentences, each randomly selected from 500 different legal cases retrieved from the Caselaw Access Project \cite{caselaw}. Second, we created a test dataset of 500 sentences, where each sentence was randomly selected from 500 different cases (also randomly selected) from the US Supreme Court cases in \cite{supremecourtcases}. Quantitative experiments such as comparison of USLT with other competitive methods are done using the sentences from the US Supreme Court cases.

\subsection{Evaluation metrics}
\label{metrics}
For the evaluation of TS methods, a common approach is to use the SARI (Simple Automated Readability Index) metric. However, SARI needs reference sentences in addition to the original sentence and the system output. Due to a lack of reference sentences in the legal domain, the SARI score cannot be used in our framework. Instead, we use statistical metrics that do not require reference sentences. Specifically, we use the Flesch–Kincaid Grade Level (FKGL) \cite{fkgradelevel} and the Dale-Chall (DC) scores \cite{dale_chall}. USLT also aims to maintain the original sentences' semantic meaning following simplification, which was assessed via a semantic difference metric. 

\textbf{Flesch–Kincaid Grade Level (FKGL):} FKGL score is a widely used readability metric, with recent adoption in legal domain \cite{fkgradelevel}. It aims to create a ``grade level" score for a text, i.e., if a text has a score 10.3, it means that the text is readable for a 10th-grade student. Hence, a lower score indicates a simpler sentence. This score is calculated with the following formula:
  \[ 
FKGL = 0.39(\frac{n_w}{n_S}) + 11.8(\frac{n_{syl}}{n_w}) - 15.59,
\]
where $n_S$ stands for the number of sentences, $n_w$ is the number of words, and $n_{syl}$ is the number of syllables \cite{kincaid1975derivation}.

\textbf{Dale-Chall Score (DC):} DC score provides a numeric assessment for the comprehension difficulty of a text. This formula is based on the usage of familiar words in English. DC score is calculated with the following formula:
  \begin{align*}
  DC &= 0.1579(\frac{n_{dw}}{n_w}) + 0.0496(\frac{n_w}{n_S}),
\end{align*}
where $n_{dw}$ stands for the number of difficult words, $n_w$ stands for the number of all words, and $n_S$ stands for the number of sentences. Similar to the FKGL score, a lower score indicates a simpler sentence \cite{dale_chall}.

\textbf{Semantic Difference (SD):} This metric measures how much the output sentences preserve the semantic content of the original sentences. To accomplish this goal, we compare vector embeddings of the original and output sentences \cite{caselaw}, which were reported to correlate with the semantic closeness \cite{shashavali2019sentence}. To eliminate unfair advantages due to syntactic changes in the output sentences, we divide both the output and input sentences into five-grams, run a sliding window through them, compute the cosine similarity between the word embeddings in two windows, and take the highest cosine similarity. Then we take the average of these similarity scores for different windows, which gives an overall similarity score $\mathrm{sim_{cos}} \in [-1,1]$.
From this score, we get a distance metric by writing:
\[
\mathrm{dist_{cos}} = 1 - \mathrm{sim_{cos}} \in [0,2].
\]
We multiply this score by $6$ to attain a range $[0,12]$, where the lower the score, the more semantically similar the output sentence is to the original. This brings the SD score to a similar scale to FKGL and DC scores, which is more suitable when calculating harmonic means in Section \ref{hyperparam_section}.

Finally, in reporting our experimental results, we perform statistical significance tests. A non-parametric Wilcoxon signed-rank test is used to compare the scores of comparison methods. 

\subsection{Hyperparameter Optimization}
\subsubsection{Optimizing the weights in the Substitution Candidate Ranking (SR) step}
\label{hyperparam_section}
As explained in Section \ref{three_step_section}, we have five weights to tune in SR step. We used Bayesian optimization over the dataset to find the most suitable weights for different features. We ran 200 iterations on a domain that included intervals between 0 and 6, i.e., $[0,6]$ for all different weights. In the process, we aim to minimize the harmonic mean of FKGL, DC, and SD scores of the outputs at each iteration for different weight combinations. We present the optimized weights in Table \ref{hyperparameter_scores}.

\begin{table}[h]
\centering
\begin{tabular}{ |p{3.7cm}||p{2.3cm}| }

 \hline
 \textbf{Hyperparameter} & \textbf{Optimized Weight}\\
 \hline
 LM probabilities ($F_B$) & $W_B = 3.00$  \\
 Cosine similarity ($F_C$) & $W_C = 1.42$  \\
 Cross-entropy loss ($F_{LM}$) & $W_{LM} = 0.36$  \\
 Frequency ($F_{F}$) & $W_F = 2.00$  \\
 Word length ($F_{L}$) & $W_L = 4.61$  \\
 \hline
\end{tabular}
\captionsetup{justification=justified}
\caption{Optimal weights used in the SR step.}
\label{hyperparameter_scores}
\end{table}

\subsubsection{Optimizing the number of suggestions}
\label{suggestion_number_optimization}
For USLT to function correctly, we should take a proper number of suggestion candidates from the SG step into the SR step. Our experimental results suggest that there is an almost monotonically decreasing relationship between the metrics with which we evaluate the complexity of sentences and the number of suggestions coming from the LM. The metrics reach a near-optimal solution at $n=76$, so we adopt this as the number of candidates to limit computational load. 

\subsection{Benchmarking}
\label{evaluations}

We quantitatively evaluate USLT and compare it with baselines using the 500 sentences randomly selected from a corpus of 27,000 US Supreme Court cases. This test dataset is different from the one on which we performed hyperparameter optimizations in Sections \ref{hyperparam_section} and \ref{suggestion_number_optimization}.

The literature has no standard baseline for simplifying legal documents since USLT is the first legal domain-specific TS method. Thus, we compare USLT by directly applying the state-of-the-art TS models, developed for daily English usage, to legal texts. We consider the following baseline models for benchmarking:

\textbf{ACCESS}: This method leverages an encoder-decoder transformer model \cite{attention} and the Wikilarge dataset \cite{DRESS}, which is a set of aligned complex-simple sentence pairs from English Wikipedia (EW) and Simple English Wikipedia (SEW) \cite{martin2020controllable}. Since we do not have such an aligned dataset in the legal domain, we cannot train a domain-specific ACCESS model but rather adopt the pre-trained version in regular language. 

\textbf{MUSS}: This method primarily relies on a single, broad corpus to detect simple and complex sentence samples and data mining to find correspondent pairs of simple-complex sentences. MUSS was originally trained on the Wikilarge corpus, and it then used ACCESS to train an LS model in a supervised fashion \cite{martin2021muss}. It is difficult to find a broad legal corpus containing simple corresponding sentences for the complex legal text. Therefore, we used the pre-trained model of MUSS as a baseline.

\textbf{REC-LS}: This TS method tries to recursively find suitable synonyms for complex words in a sentence one by one. REC-LS utilizes an online thesaurus that provides simpler alternatives for different words \cite{gooding-kochmar-2019-recursive}.

\textbf{LS-BERT}: This is an LS method utilizing BERT \cite{qiang21lsbert}. LS-BERT leverages the first three features contained in USLT for candidate ranking, without weighting, and a domain-general CWI step. 

\textbf{LS-BERT\textsubscript{CWI}} (LS-BERT enhanced with the CWI system of USLT): To eliminate the possible advantage of our CWI component, we let LS-BERT use our CWI.

\section{Results}
\label{results}

\subsection{Quantitative results}
\begin{figure*}[h]
\centering
\adjustimage{width=\textwidth, center}{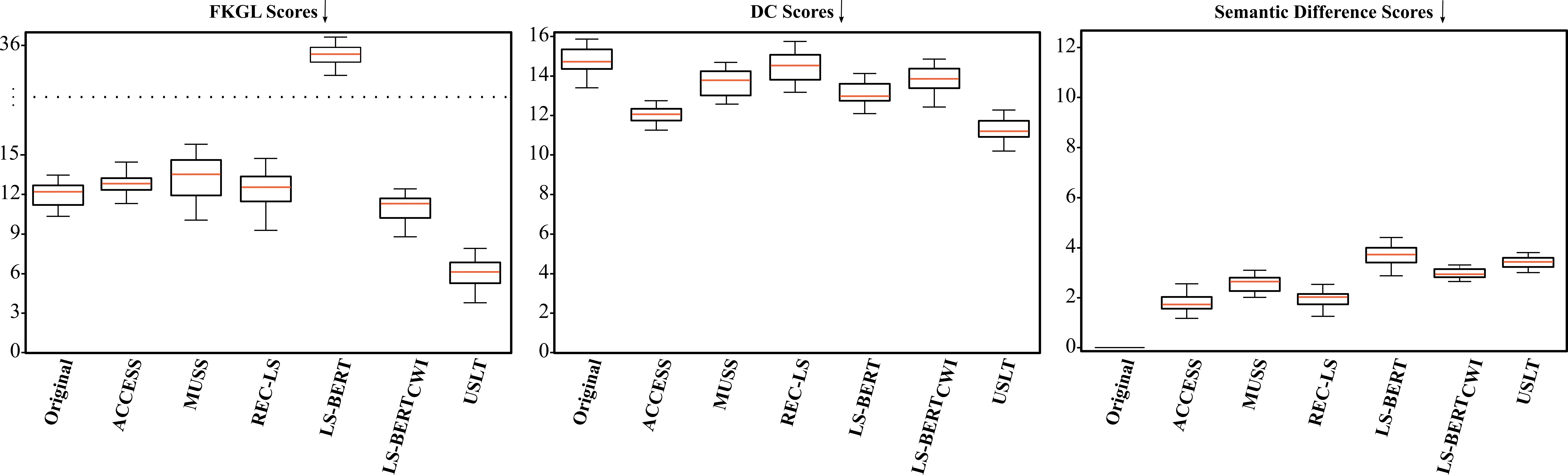}
\captionsetup{justification=justified}
\caption{Comparison of our model against the baseline models. In all metrics, a lower score is desired. The proposed method, USLT, achieves the lowest score in the first two plots, indicating that we achieve the simplest sentences in terms of readability. In the last plot, we show that USLT is approximately on par with the other methods in the semantic coherence of the output sentences.}
\label{metric_comparison}
\end{figure*}

The performance of USLT on simplifying legal texts is compared with the baseline models ACCESS, MUSS, REC-LS, LS-BERT, and LS-BERT$\mathrm{_{CWI}}$, based on the FKGL, DC, and SD metrics. To run these experiments, we used 500 randomly chosen sentences (see Section \ref{datasets}). We randomly group the chosen sentences into 10 chunks, each consisting of 50 sentences. Then, we run the baseline models in Section \ref{evaluations} and the USLT to calculate FKGL, DC, and SD scores. Results are reported as box-plots in Fig. \ref{metric_comparison}. Notice that in Fig. \ref{metric_comparison}, for the first two metrics that measure simplicity, USLT obtains the most favorable results. In the last metric, we show that USLT is on par with other methods on semantic preservation. Thus, USLT is competitive in retaining the semantic contents of the input text. Naturally, methods that are conservative and thereby limited in simplification, such as REC-LS can yield higher SD scores \cite{qiang21lsbert}.

We carried out statistical significance tests on our results. We obtained p-values for all models less than $0.05$ (for most of them, even less than $0.005$) so that the reported results are statistically significant. These statistical significance tests are presented in Table \ref{p_value_othermethods}.

\begin{table}[h]
\centering
\begin{tabular}{ |p{3.24cm}|| p{1.60cm} | p{1.22cm} | }
 \hline
 \textbf{First Model/ Second Model} & \textbf{FKGL} & \textbf{DC}\\
 \hline
 USLT/ Original Text & <0.005 & <0.005 \\
 USLT/ ACCESS & <0.005 & <0.005 \\
 USLT/ MUSS & <0.005 & <0.005 \\
 USLT/ REC-LS & <0.005 & <0.005\\
 USLT/ LS-Bert  & <0.005 & 0.048 \\
 USLT/ \texorpdfstring{LS-Bert\textsubscript{CWI}}  & 0.009 & 0.042 \\
 \hline
\end{tabular}
\captionsetup{justification=justified}
\caption{Significance testing results for comparisons among models. A p-value less than 0.05 indicates that the first model significantly outperforms the second model in the respective quantitative metric. }
\label{p_value_othermethods}
\end{table}

\subsection{Qualitative assessment}

We also inspect simplified sentences in the context of original sentences to qualitatively assess the success of various methods. Representative qualitative examples from USLT and baseline models are provided in Table \ref{qualitative_comparison}. When a particular word or phrase (i.e., the Latin phrases used commonly in the legal domain, such as \verb|bona fide| or \verb|actus reus|), USLT detects the complicated word and replaces it with a suitable substitution, when there are multiple words of this nature, USLT can also successfully find simpler replacements sequentially.

\begin{table*}[h]
\centering
\begin{tabularx}{\textwidth}{c>{\raggedright}X}
\toprule
\textbf{Model Name} & \textbf{Sentence} \tabularnewline
\midrule
\emph{Original} & It is the mental state of mind of the person at the time the actus reus was committed. \tabularnewline

\emph{ACCESS} & It is the mental state of mind of the person at the time the actus was  \textbf{sent}. \tabularnewline

\emph{MUSS} & It is the mental state of mind of the person at the time the actus reus was committed. \tabularnewline

\emph{REC-LS} &It is the mental state of mind of the person at the time the actus reus was committed. \tabularnewline

\emph{LS-BERT} &  It is the \textbf{actual} state of mind of the person at the time the \textbf{alleged act} was \textbf{done}. \tabularnewline

\emph{USLT} & It is the mental state of mind of the person at the time. The \textbf{act} was committed the time. \tabularnewline

\midrule
\emph{Original} & Before filing a petition for a divorce the plaintiff must have lived within the state at least one year. \tabularnewline

\emph{ACCESS} &Before filing a petition for a divorce the plaintiff must have lived within the state at least one year. \tabularnewline

\emph{MUSS} & The plaintiff must have lived in the state for at least one year before filing a divorce petition. \tabularnewline

\emph{REC-LS} & Before filing a petition for a \textbf{divorcement} the \textbf{litigant} must have lived within the state at least one year. \tabularnewline

\emph{LS-BERT} & Before bringing a \textbf{complaint} for a \textbf{judgment} the \textbf{defendant} must have lived within the state at least one year. \tabularnewline

\emph{USLT} & The \textbf{spouse} must have lived within the state at least one year. The \textbf{spouse} was filing a \textbf{bill} for a divorce. \tabularnewline

\midrule
\emph{Original} & We cannot give you away because for us, you are indispensable to prosecute the plaintiff who committed the actus reus. \tabularnewline

\emph{ACCESS} & We cannot give you away because \textbf{they are not allowed} to prosecute the plaintiff who committed the actus reus. \tabularnewline

\emph{MUSS} & We cannot give you away because for us, you are essential to bring the actus reus case \textbf{against the defendant}. \tabularnewline

\emph{REC-LS} & We can not give you away because for us, you are \textbf{vital} to prosecute the \textbf{litigant} who \textbf{perpetrated} the actus reus. \tabularnewline

\emph{LS-BERT} & We can not give you away because for us, you are \textbf{going to be the person} who is the actus \textbf{se}. \tabularnewline

\emph{USLT} & We can not give you away. You are \textbf{necessary} to prosecute the \textbf{defendant}. The \textbf{defendant} committed the \textbf{crime}. \tabularnewline
\bottomrule

\end{tabularx}
\captionsetup{justification=justified}
\caption{In the first example, contrary to other methods, USLT can identify ``actus reus" as a complex phrase and substitute ``act" instead. In the second example, USLT can identify and simplify multiple complex words. In the third example, USLT can identify multiple complex words and complex phrases in a given sentence and find simpler alternatives for them. In all examples, USLT also successfully carries out sentence splitting contrary to the baselines.}
\label{qualitative_comparison}
\end{table*}

\subsection{Ablation Studies}
Ablation studies are carried out to characterize the importance of different components in USLT. The results are presented in Table \ref{ablation_table}. We first inspected which components are vital for the LS stage of the algorithm. In the experiments, we see that the usage of the features $F_B$, $F_C$, and $F_{LM}$ are important to provide contextual information to the method and mainly affect the SD score. On the other hand, frequency and length features help USLT select shorter and simpler words. Thus, combining these features allows USLT to maintain simplicity and semantic coherence in the output sentences. Furthermore, we observe that applying the sentence-splitting algorithm causes an increase in the FKGL and DC scores, suggesting improved comprehensibility. This comes at the price of a modest decrease in SD scores, which is expected since the decomposition of a long sentence inevitably involves some degree of syntactic and structural changes. 

\begin{table}[h]

\centering
\begin{tabular}{ |p{3.63cm}||p{1.10cm}|p{0.70cm}|p{0.70cm}| }

 \hline
 \textbf{Model Name} & \textbf{FKGL $\downarrow$} & \textbf{DC $\downarrow$} & \textbf{SD $\downarrow$} \\
 \hline
 
 LS w/o BERT prob. feature & 10.95 & 13.90 & 0.50  \\
 LS w/o cosine similarity feature & 10.95 & 13.90 & 0.50 \\
 LS w/o LM loss feature & 11.04 & 13.90 & 0.50 \\
 LS w/o frequency feature & 11.83 & 14.52 & 0.40 \\
 LS w/o word length feature & 10.96 & 13.90 & 0.49 \\
 LS w/o sentence splitting & 10.95 & 13.90 & 0.49 \\
 LS w/ sentence splitting (USLT) & 2.35 & 12.23 & 1.84 \\
\hline
\end{tabular}
\captionsetup{justification=justified}
\caption{Results of ablation studies for various models in terms of FKGL, DC, and, SD scores.}
\label{ablation_table}
\end{table}

\section{Conclusion}
\label{conclusion}
In this study, we introduced an unsupervised method to simplify legal texts through both lexical-level and syntactic-level alterations. Our proposed approach first identifies the complex words in a given sentence through an analysis of the frequency of occurrence of words and uses relevant measures such as Zipf Scale to quantify the complexity levels. Substitution candidates for complex words are then generated using masked-out word prediction in transformer-based language models. Candidates are ranked according to several word features, and the highest ranking candidate is selected for replacement. Finally, sentence splitting is performed to decompose prolonged sentences into shorter ones to improve legibility. 

Experimental results show that the proposed method provides advantages over the previous domain-general methods devised for regular language. Using domain-specific corpora and language models in USLT helps improve performance in simplifying legal texts. As USLT adopts state-of-the-art domain-specific transformer models as an LM backbone, its outputs are expected to be compatible with various high-level legal NLP tasks. 

\ifCLASSOPTIONcaptionsoff
  \newpage
\fi

\balance
\bibliographystyle{IEEEtran}
\bibliography{references_abbrv}
\end{document}